\documentclass[11pt, a4paper, logo, copyright, nonumbering]{deepseek}
\usepackage[numbers, sort&compress]{natbib}
\definecolor{citecolor}{HTML}{1976D2}
\hypersetup{
    colorlinks=true,
    linkcolor=black,
    filecolor=magenta,
    urlcolor=cyan,
    citecolor=citecolor,
}
\usepackage{dblfloatfix}
\usepackage{ulem}
\usepackage{caption}
\usepackage{dramatist}
\usepackage{xspace}
\usepackage{pifont}
\usepackage{multirow}
\usepackage{tcolorbox}
\usepackage{xltabular}
\usepackage{longtable}
\usepackage{hyperref}
\usepackage{amsfonts}
\usepackage{amsmath}
\usepackage{amssymb}
\usepackage{lineno}

\usepackage[bottom]{footmisc}

\usepackage{CJKutf8}
\usepackage{subfigure}
\usepackage{setspace}
\usepackage{xcolor}

\newlength\savewidth\newcommand\shline{\noalign{\global\savewidth\arrayrulewidth
  \global\arrayrulewidth 1pt}\hline\noalign{\global\arrayrulewidth\savewidth}}

\makeatletter
\def\@BTrule[#1]{%
  \ifx\longtable\undefined
    \let\@BTswitch\@BTnormal
  \else\ifx\hline\LT@hline
    \nobreak
    \let\@BTswitch\@BLTrule
  \else
     \let\@BTswitch\@BTnormal
  \fi\fi
  \global\@thisrulewidth=#1\relax
  \ifnum\@thisruleclass=\tw@\vskip\@aboverulesep\else
  \ifnum\@lastruleclass=\z@\vskip\@aboverulesep\else
  \ifnum\@lastruleclass=\@ne\vskip\doublerulesep\fi\fi\fi
  \@BTswitch}
\makeatother

\addto\extrasenglish{
}

 {\begin{list}{}%
         {\setlength{\leftmargin}{#1}}%
         \item[]%
 }
 {\end{list}}
 
\reportnumber{001} 

\renewcommand{\today}{}

\title{\centering DeepSeek-VL2: Mixture-of-Experts 
Vision-Language Models for Advanced Multimodal Understanding}

\author[*]{
\small
Zhiyu Wu$^{*}$, 
Xiaokang Chen$^*$, 
Zizheng Pan$^*$, 
Xingchao Liu$^*$, 
Wen Liu$^{*, \dag}$,  
Damai Dai,
Huazuo Gao,\quad
Yiyang Ma, 
Chengyue Wu, 
Bingxuan Wang, 
Zhenda Xie, 
Yu Wu, 
Kai Hu,
Jiawei Wang, 
Yaofeng Sun, 
Yukun Li, 
Yishi Piao, 
Kang Guan, 
Aixin Liu, 
Xin Xie, 
Yuxiang You, 
Kai Dong, 
Xingkai Yu, 
Haowei Zhang, 
Liang Zhao, 
Yisong Wang, 
Chong Ruan$^\ddag$ \\

\small
DeepSeek-AI

}
\correspondingauthor{\footnotesize\, 
$^*$: Core contributors.  $^{\dag}$: Project lead.
$^\ddag$: Corresponding author.
}

\newcommand{\eg}{\textit{e}.\textit{g}.}

\newcommand{\codename}{DeepSeek-VL2}

\begin{abstract}
We present DeepSeek-VL2, an advanced series of large Mixture-of-Experts (MoE) Vision-Language Models that significantly improves upon its predecessor, DeepSeek-VL, through two key major upgrades. 
For the vision component, we incorporate a dynamic tiling vision encoding strategy designed for processing high-resolution images with different aspect ratios.
For the language component, we leverage DeepSeekMoE models with the Multi-head Latent Attention mechanism, which compresses Key-Value cache into latent vectors, to enable efficient inference and high throughput.
Trained on an improved vision-language dataset, DeepSeek-VL2 demonstrates superior capabilities across various tasks, including but not limited to visual question answering, optical character recognition,  document/table/chart understanding, and visual grounding. Our model series is composed of three variants: DeepSeek-VL2-Tiny, DeepSeek-VL2-Small and DeepSeek-VL2, with 1.0B, 2.8B and 4.5B activated parameters respectively.
DeepSeek-VL2 achieves competitive or state-of-the-art performance with similar or fewer activated parameters compared to existing open-source dense and MoE-based models.
Codes and pre-trained models are publicly accessible at \url{https://github.com/deepseek-ai/DeepSeek-VL2}.

\end{abstract}

\begin{document}
\begin{CJK*}{UTF8}{gbsn}

\maketitle

\begin{figure}[ht]
	\centering
    \includegraphics[width=0.54\linewidth]{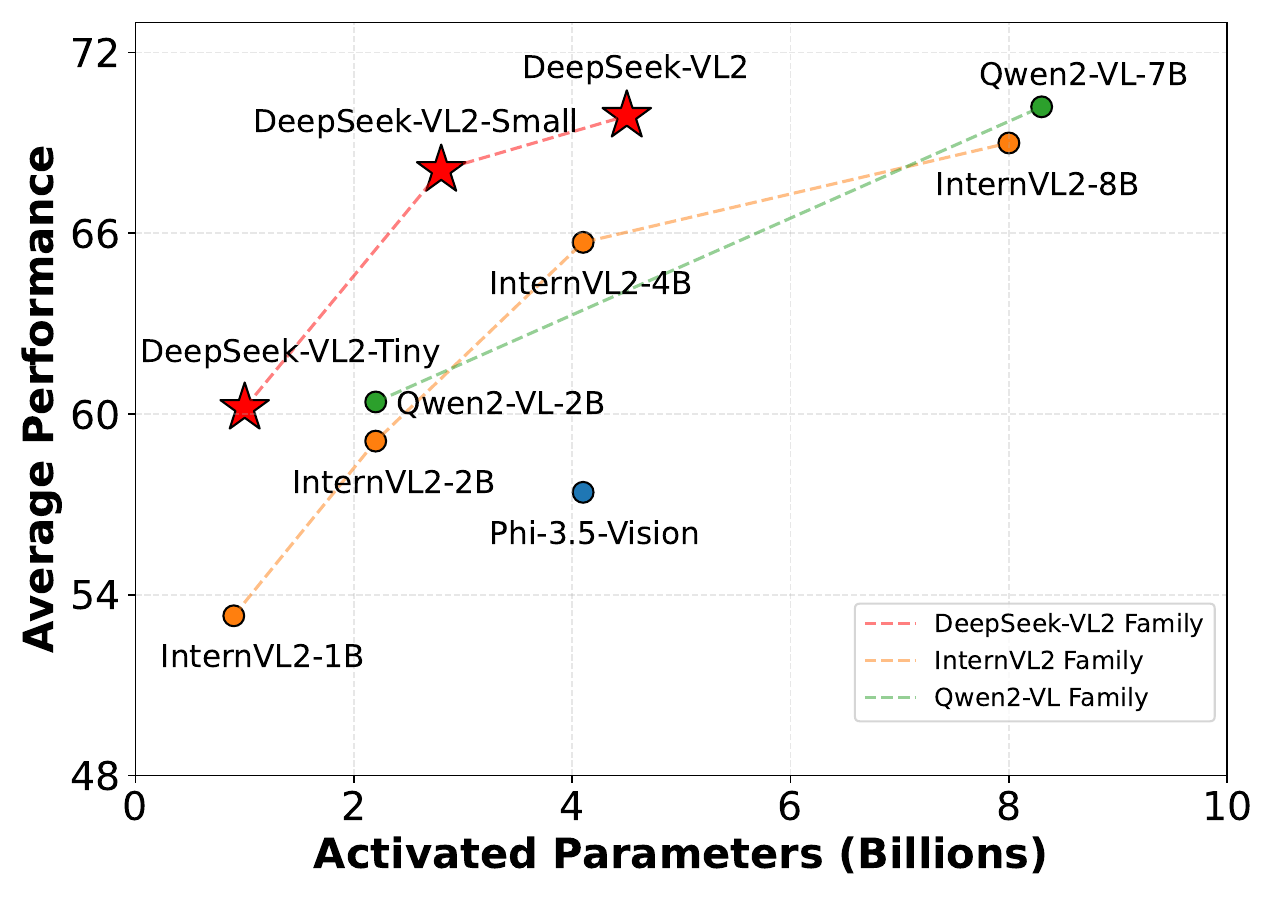}
    \vspace{-2mm}
	\caption{\textbf{Average performance vs. activated parameters among different open-source models}. We average the accuracy of MMBench v1.1, MMStar, MMMU (Val), MathVista (TestMini), AI2D (Test), and OCRBench. The scores of OCRBench are divided by $10$ to scale them to $[0, 100]$.}
	\label{fig:teaser}
\end{figure}

\newpage

\begin{spacing}{0.9}
\tableofcontents
\end{spacing}

\newpage

\section{Introduction}
Large Vision-Language Models (VLMs) have emerged as a transformative force in artificial intelligence~\cite{liu2024visual,lu2024deepseek,wu2024janus,ma2024janusflow, tongcambrian, chen2023internvl, wang2024qwen2}, extending the remarkable capabilities of Large Language Models (LLMs) to seamlessly process both visual and textual information. This advancement has dramatically expanded the potential for AI systems to tackle complex real-world applications that require multimodal understanding.

In this technical report, we present DeepSeek-VL2, a new series of open-source Vision-Language Models that leverages the Mixture-of-Experts (MoE) architecture to achieve substantial improvements in both performance and efficiency compared to its predecessor, DeepSeek-VL~\citep{lu2024deepseek}. Our advancements center around three key aspects: (1) a dynamic, high-resolution vision encoding strategy that enhances visual understanding, (2) an optimized language model architecture that significantly improves both training and inference efficiency, and (3) a refined vision-language data construction pipeline that not only boosts overall performance but also extends model capabilities to new areas such as precise visual grounding.

For the vision component, we introduce a dynamic tiling vision encoding strategy that efficiently processes high-resolution images of varying aspect ratios. This approach improves over DeepSeek-VL's hybrid vision encoder, which extracted features from images at two fixed resolutions ($384 \times 384$ and $1024 \times 1024$).
Our approach avoids the limitations of the old fixed-size encoder and excels in tasks requiring ultra-high resolution, including visual grounding, document/table/chart analysis, and detailed feature extraction, while maintaining a manageable number of visual tokens. Drawing inspiration from established slicing-tile methods, our system dynamically segments high-resolution inputs into local tiles, processes each tile through a shared vision transformer, and seamlessly integrates the extracted features within the language model. This design preserves the advantages of vision transformers with local attention, enabling rich feature extraction without the quadratic computational scaling typically associated with increasing image resolutions.

For the language component, we leverage DeepSeek language models~\citep{dai2024deepseekmoe, ds_v2}, featuring the Multi-head Latent Attention (MLA) mechanism. MLA significantly reduces computational cost by compressing the Key-Value (KV) cache into a latent vector, resulting in faster inference and increased throughput capacity. We further enhance efficiency through the DeepSeekMoE framework~\citep{dai2024deepseekmoe, noaux_tc}, which employs sparse computation techniques. 
Our model series adopt three MoE variants, 3B, 16B, and 27B. These LLMs have 0.57B, 2.4B, and 4.1B activated parameters respectively.

We also greatly enhance our vision-language training data in terms of quality, quantity, and diversity. This comprehensive dataset enables better generalization and performance across a broad spectrum of tasks, including Visual Question Answering (VQA), Optical Character Recognition (OCR), document/table/chart understanding, visual reasoning, and general chatbot applications. The improved training data has also enabled new abilities such as visual grounding and Graphical User Interface (GUI) perception.

In summary, DeepSeek-VL2 marks a substantial leap forward in large-scale Mixture-of-Experts Vision-Language modeling. Through a new visual processing strategy and an optimized language model, we develop a series of models that balances performance with efficiency. By open-sourcing the pre-trained models, we aim to accelerate progress in the field and promote collaborative research advancement.

\begin{figure}[!ht]
	\centering
	\includegraphics[width=\linewidth]{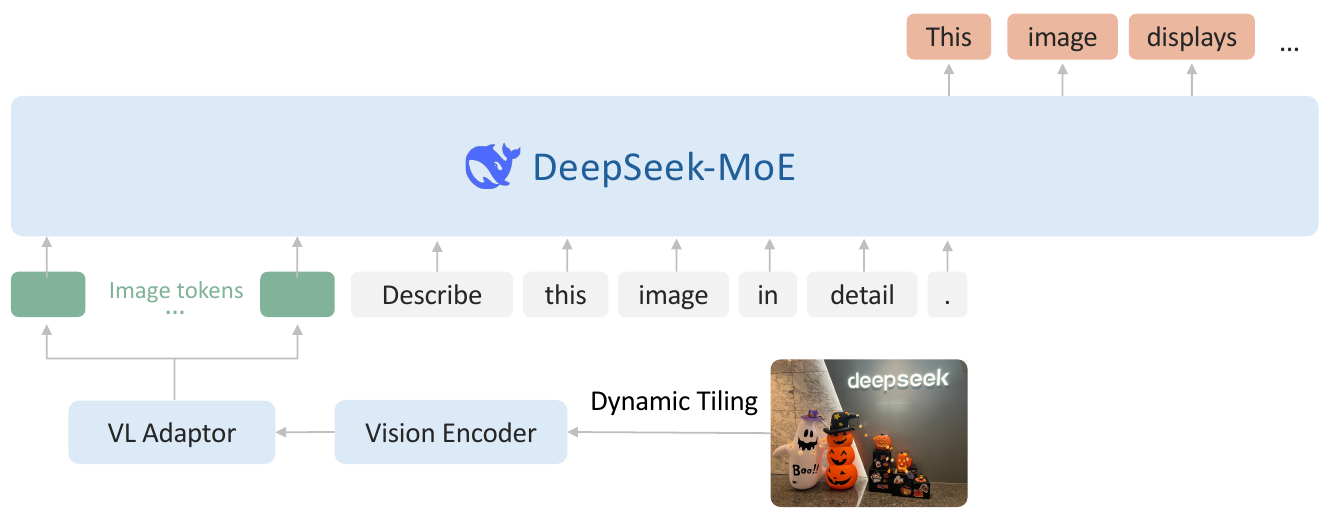}
	\caption{\textbf{Overview of \codename}. The overall structure is a llava-style architecture, which includes a vision encoder, a VL adaptor, and a MoE-based LLM.}
	\label{fig:main_framework}
\end{figure}

\section{Model Architecture}

DeepSeek-VL2 consists of three core modules: (1) a vision encoder, (2) a vision-language adaptor, and (3) a Mixture-of-Experts language model. Building upon the decoder-only LLaVA-style~\citep{liu2024visual} architecture of its predecessor, DeepSeek-VL2 introduces two major advancements: a dynamic tiling strategy and a DeepSeekMOE~\citep{dai2024deepseekmoe, noaux_tc} language model featuring Multi-head Latent Attention~\citep{ds_v2}. These innovations enable more efficient processing of both high-resolution visual inputs and text data.

\textbf{Dynamic Tiling Strategy.} The original DeepSeek-VL employed a hybrid vision encoder combining SigLIP~\citep{zhai2023sigmoid} for coarse-grained feature extraction at $384 \times 384$ resolution and SAM-B~\citep{kirillov2023segment} for fine-grained feature extraction at $1024 \times 1024$ resolution. While this fusion approach generated rich visual representations suitable for various vision-language tasks, it was limited by the fixed $1024 \times 1024$ resolution constraint. This limitation is particularly challenging for processing images with larger resolutions and extreme aspect ratios, such as those found in InfographicVQA~\citep{mathew2022infographicvqa}, dense OCR, and detailed visual grounding tasks.

Inspired by recent advances in VLMs~\citep{liu2024llavanext, nvlm2024, chen2024internvl2}, we implement a dynamic tiling strategy by splitting a high-resolution image into tiles. This approach enables the efficient processing of different high-resolution images with varying aspect ratios using a single SigLIP-SO400M-384 vision encoder~\citep{zhai2023sigmoid}.
The pre-trained SigLIP operates at a base resolution of $384 \times 384$. 
To accommodate different aspect ratios, we define a set of candidate resolutions:
$C_R = \{(m \cdot 384, n \cdot 384)~\mid~m \in \mathbb{N}, n \in \mathbb{N},1\le m,n,mn\le9\}$, 
where $m:n$ represents the aspect ratio. 
For an input image of size $(H, W)$, we calculate the padding area required for resizing\footnote{We first resize the original image until its long side matches the target resolution, then pad the other dimension while maintaining the original aspect ratio.} it to each candidate resolution in $C_R$. We select the resolution $(m_i \cdot 384, n_i \cdot 384)$ that minimizes the padding area. The resized image is then divided into $m_i \times n_i$ local tiles of $384 \times 384$ pixels, plus one global thumbnail tile. The SigLIP-SO400M-384 vision encoder processes all $(1+ m_i \times n_i)$ tiles, yielding $27 \times 27 = 729$ visual embeddings of $1152$ dimensions per tile. 
For computational efficiency and context length management, we disable the dynamic tiling strategy when processing multiple $(>2)$ images.

\begin{figure}[!ht]
	\centering
	\includegraphics[width=0.99\linewidth]{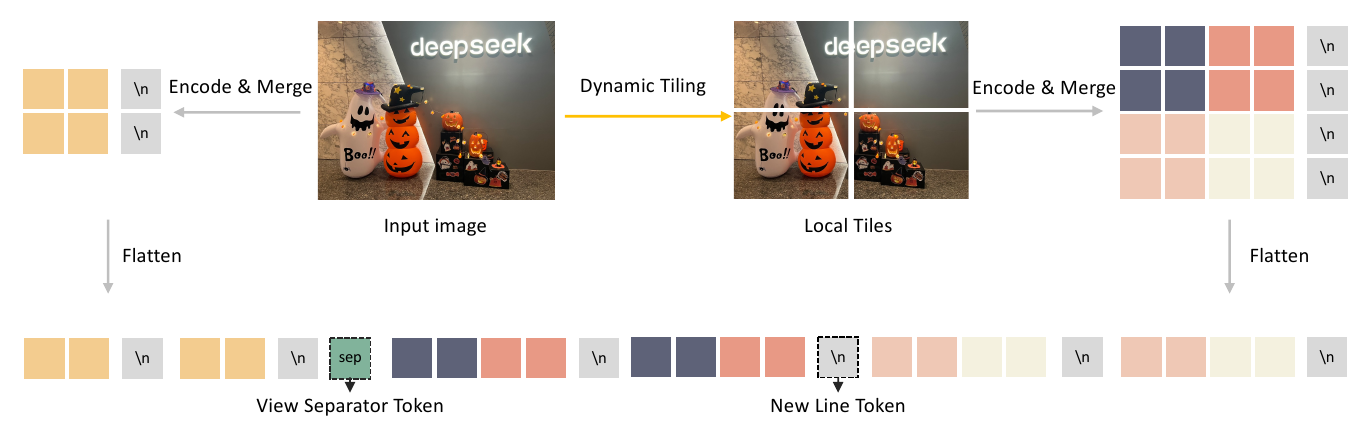}
	\caption{\textbf{Illustration of dynamic tiling strategy in DeepSeek-VL2}. By dividing images into multiple tiles, DeepSeek-VL2 achieves stronger fine-grained understanding capabilities compared to DeepSeek-VL.
}
	\label{fig:image_crop_dsvl_2}
\end{figure}

\textbf{Vision-Language Adaptor.} 
Following visual tile processing, we implement a $2 \times 2$ pixel shuffle operation to compress each tile's visual tokens from $27 \times 27$ to $14 \times 14=196$ tokens. We then introduce three special tokens when processing the $(1 + m_i \times n_i)$ tiles. For the global thumbnail tile ($14 \times 14$), we add 14 \texttt{<tile\_newline>} tokens to the end of each row, resulting in a total number of $14 \times 15 = 210$ tokens. For the $m_i \times n_i$ local tiles, which are arranged in a 2D grid of shape $(m_i \cdot 14, n_i \cdot 14)$, we append $m_i \cdot 14$ \texttt{<tile\_newline>} tokens at the end of the final column to indicate the end of a row of all the local tiles. Additionally, a \texttt{<view\_separator>} token is inserted between the global thumbnail tile and the local tiles. The complete visual sequence contains $210 + 1 + m_i \cdot 14 \times (n_i \cdot 14 + 1)$ visual tokens, which are subsequently projected into the language model's embedding space using a two-layer multilayer perceptron (MLP). A visual illustration of our dynamic tiling strategy is shown in Figure~\ref{fig:image_crop_dsvl_2}.

\textbf{DeepSeekMoE LLM.} 
Our language model is based on DeepSeekMoE~\citep{dai2024deepseekmoe, noaux_tc}, which incorporates the Multi-head Latent Attention mechanism~\citep{ds_v2}. MLA enhances inference efficiency by compressing the Key-Value cache into a latent vector, enabling increased throughput capacity.
The model also incorporates a MoE architecture~\cite{dai2024deepseekmoe} allowing for efficient inference through sparse computation. During MoE training, we introduce a global bias term~\cite{noaux_tc} for each expert to cost-effectively improve load balancing between experts. DeepSeek-VL2 comes in three variants with the following model sizes: 1.0B, 2.8B and 4.5B. Complete architectural specifications can be found in Table~\ref{tab:arch}.

\begin{table}[t!]
\caption{
\textbf{Architectural configuration for DeepSeek-VL2}. We list the hyperparameters of the architecture along with the details related to the mixture-of-expert training.
}
\centering
\renewcommand{\arraystretch}{1.2}
\scalebox{0.8}{
\begin{tabular}{l|c|c|c}
    \shline
    & \textbf{DeepSeek-VL2-Tiny} & \textbf{DeepSeek-VL2-Small} & \textbf{DeepSeek-VL2} \\
    \shline
    Vocabulary size & 129,280 & 102,400 & 129,280 \\
    Embedding size & 1,280 & 2,048 & 2,560 \\
    \#Attention heads & 10 & 16 & 32 \\
    \#Layers & 12 & 27 & 30 \\
    Attention & Multi-Head Attention & MLA (rank=512) & MLA (rank=512) \\
    \#Routed experts & 64 & 64 & 72 \\
    \#Shared experts & 2 & 2 & 2 \\
    Top-K for expert selection & 6 & 6 & 6 \\
    Routing function & Softmax & Softmax & Sigmoid \\
    Expert correction bias & $\times$ & $\times$& \checkmark \\
    \shline
    
\end{tabular}
}
\label{tab:arch}
\end{table}
\section{Data Construction}
We build a comprehensive Vision-Language dataset from diverse sources for DeepSeek-VL2.  
The training process is structured into three distinct stages: (1) VL alignment, (2) VL pretraining, and (3) supervised fine-tuning (SFT). In the following parts, we provide descriptions of the data used in each stage.

\subsection{Vision-Language Alignment Data}
\label{data:vl_adaptor}
The alignment stage focuses on training the MLP connector to bridge the pretrained visual encoder and the LLM. For this initial warmup phase, we utilize ShareGPT4V~\cite{sharegpt4v}, a dataset containing approximately 1.2M caption and conversation samples.

\subsection{Vision-Language Pretraining Data}
\label{data:vl_pretraining}
Following DeepSeek-VL~\cite{lu2024deepseek}, our pretraining data combines vision-language (VL) and text-only data to maintain a balance between VL capabilities and text-only performance. For DeepSeek-VL2, we maintain a ratio of around 70\% VL data to 30\% text-only data, with the latter sourced directly from our base LLM pretraining corpus. In the following, we categorize the VL data into several groups and describe their details.

\paragraph{Interleaved image-text data.} 
Our data collection begins with several open-sourced datasets, including WIT~\cite{wit}, WikiHow~\cite{wikihow}, and 30\% random samples from OBELICS~\cite{obelics}. This specific mixing ratio was determined through preliminary experiments with DeepSeek-VL2-Tiny. To enhance multilingual capabilities, we supplemented the predominantly English datasets with Chinese content extracted from Wanjuan~\cite{wanjuan}. Additionally, we developed an in-house collection to expand coverage of general real-world knowledge.

\paragraph{Image captioning data.} 
Image captions represent fundamental data in VLM training, providing direct alignment between visual and textual information. We initially leveraged diverse open-source datasets \cite{densefusion, pixelprose, wikiart, li2024mmsci, arxivcap, yfcc100m, laion_aesthetics, sam_dataset, OpenImages, megalith10m_dataset, JourneyDB, pick_a_pic, dalle3_1m, wukong}. However, our preliminary analysis revealed severe quality variations across these datasets, ranging from dense, accurate captions generated by advanced VLMs to problematic cases with brief descriptions, mismatched text pairs, or obvious hallucinations.
To address these quality inconsistencies, we developed a comprehensive image captioning pipeline that considers: (1) OCR hints, (2) meta information (e.g., location, camera settings), and (3) relevant original captions as prompts. Using an in-house captioner, we recaption the images following prompting strategies similar to PixelProse~\cite{pixelprose}, employing varied instructions to guide the VLM's caption generation.

Despite the overall improvement in caption quality, we observed repetition issues in the large-scale annotation pipelines. To mitigate this, we implemented a quality control pipeline using DeepSeek Chat~\cite{ds_v2} to score all captions simply based on their writing quality. In practice, this approach is both efficient and effective in filtering out low-quality captions.

\paragraph{Optical character recognition data.}
To develop OCR capabilities, we used open-source datasets including LaTeX OCR~\cite{latexocr} and 12M RenderedText~\cite{renderedtextdataset}. We combined these datasets with an extensive in-house OCR dataset covering diverse document types. Currently, our in-house dataset mainly focuses on English and Chinese character recognition. We plan to expand to other languages in our future work.

\paragraph{Visual question-answering (QA) data.}
In our early exploration, we found general QA data clearly benefits model pretraining. 
Consequently, we developed a comprehensive visual QA dataset consisting of the following categories:

\begin{itemize}[itemsep=1ex]
  \item \textbf{General VQA.} We inherit the general VQA data from DeepSeek-VL. For more details, please refer to~\citep{lu2024deepseek}.
  \item \textbf{Table, chart and document understanding.} We adopt PubTabNet~\cite{PubTabNet}, FinTabNet~\cite{FinTabNet} and Docmatix~\cite{docmatix} to enhance document comprehension capabilities.
  \item \textbf{Web-to-code and plot-to-Python generation.} We leverage Websight~\cite{websight} for webpage-to-code abilities and Python plots obtained from public Jupyter notebooks, following DeepSeek-VL. We enhance this dataset by replicating a portion of Websight using DeepSeek V2.5. We also exploit Python plot codes generated by DeepSeek V2.5 to mitigate the noises in the plot-to-code data.
  \item\textbf{QA with visual prompt.} We follow~\cite{vip_llava} to construct visual prompt understanding data by overlaying various visual indicators (arrows, boxes, circles, and scribbles) onto images from~\citep{vip_llava,as_core,wang2025all}. We then created QA pairs focusing on objects highlighted by these visual prompts.
\end{itemize}

\paragraph{Visual grounding data.} We construct our visual grounding dataset from \cite{peng2023kosmos2,shao2019objects365}. For each image's object detection annotations, we structure the data as follows:

\begin{itemize}
    \item Prompt: \texttt{Locate <|ref|><query><|/ref|> in the given image.}
    \item Response: \texttt{<|ref|><query><|/ref|><|det|>[[x1, y1, x2, y2],\ldots]<|/det|>}
\end{itemize}

\noindent during training, the question prompts are randomly sampled from a candidate pool during training. \texttt{<|ref|>}, \texttt{<|/ref|>}, \texttt{<|det|>}, \texttt{<|/det|>} are special tokens. \texttt{<query>} is a placeholder for either the category name (e.g., ``car'') or description of the object (e.g., ``the leftmost person''). \texttt{[[x1, y1, x2, y2], \ldots]} is a list of bounding boxes, where each bounding box corresponds to an object's position. 
The coordinates \texttt{x1, y1} and \texttt{x2, y2} specify the top-left and bottom-right corners respectively, normalized to values between 0 and 999 according to the resolution of the image.
We also construct negative samples where queried objects are intentionally absent from the images to enhance the robustness of the model.

\paragraph{Grounded conversation data.} 
We derived our grounded conversation dataset from \cite{peng2023kosmos2}, structured in the following format:

\begin{itemize}
    \item Prompt: \texttt{<|grounding|>Can you describe the content of the image?}
    \item Response: \texttt{Two <|ref|>dogs<|/ref|><|det|>[[x1, y1, x2, y2],\ldots]<|/det|> are running on the grass.}
\end{itemize}

\noindent As in other visual grounding data, \texttt{<|grounding|>}, \texttt{<|ref|>}, \texttt{<|/ref|>}, \texttt{<|det|>}, \texttt{<|/det|>} are special tokens and \texttt{x1, y1, x2, y2} is subject to the same normalization scheme.

\subsection{Supervised Fine-tuning Data}
\label{data:vl_sft}
Our SFT data combines a diverse collection of open-sourced datasets with high-quality in-house QA pairs. Below, we detail our efforts to enhance the quality of our SFT dataset.

\paragraph{General visual question-answering.} 
While public visual QA datasets are diverse~\citep{shah2019kvqa, chen2024allava, laurencon2024matters, vip_llava, balanced_vqa_v2, hudson2019gqa, li2024llavanext}, they often suffer from three main limitations: (1) short responses, (2) poor OCR quality, and (3) hallucinated content. To address these issues, we regenerate responses by jointly considering the original questions, images, and OCR information. Our experiments demonstrate that this approach produces more comprehensive and accurate results. During development, we observed that an early version of DeepSeek-VL2, particularly the Tiny variant, occasionally inserted English words inappropriately in Chinese responses. This issue was not present in our larger models, suggesting it stemmed from limited model capacity and an imbalance between English and Chinese data in the visual-language pretraining stage. To address this limitation in our smaller model, we developed an in-house Chinese QA dataset with diverse image descriptions and single/multi-round conversations. This dataset helps to mitigate the language mixing issue.
Furthermore, we created an extra in-house dataset to complement real-world and cultural visual knowledge, including anime, memes, cuisine and art.

\paragraph{OCR and document understanding.}
Thanks to our advanced image captioning pipeline, DeepSeek-VL2 already demonstrates superior OCR capabilities compared to other state-of-the-art VLMs. Therefore, rather than further enhancing OCR performance during the SFT stage, we focused on cleaning existing open-source datasets~\citep{laurencon2024matters, diem2014icfhr, chromewriting, yuan2022syntax, mathew2021docvqa, singh2019towards, hudson2019gqa, mathew2022infographicvqa} by removing samples with poor OCR quality.
For document understanding, we curated a diverse subset of document pages from our in-house data. We then generate multi-round conversational QA pairs specific to document comprehension. Early results indicate that this approach improves document-based interactions.

\paragraph{Table and chart understanding.}
We enhanced table-based QA data by regenerating responses for all public datasets~\citep{chentabfact, li2024multimodal} based on their original questions except Cauldron~\cite{laurencon2024matters}, which already exhibits high quality. Similar to our OCR capabilities developed during VL pretraining, our model demonstrated strong performance in chart understanding without requiring additional efforts.

\paragraph{Reasoning, logic, and mathematics.}
We enhance public reasoning-focused datasets~\citep{shi2024math, laurencon2024matters, lu2021iconqa, cherian2022deep, yu2023metamath, zhang2024mavismathematicalvisualinstruction} with more detailed reasoning processes and standardize response formats which puts the final answer at the end of the response. We observe that detailed responses are less effective when training smaller VLMs. In our exploration, DeepSeek-VL2-Tiny shows better performance with more concise responses.

\paragraph{Textbook and academic questions.}
We build an internal dataset focused on textbooks from our document collection. This dataset primarily emphasizes college-level contents across multiple academic disciplines.

\paragraph{Web-to-code and plot-to-Python generation.} We expand our in-house dataset for web code and Python plot code beyond what was used during pretraining. For open-source datasets, we improve their quality by regenerating their answers.

\paragraph{Visual grounding.}
We develop our visual grounding dataset using data from \cite{wang2023v3det,mao2016generation,yu2016modeling,wave_ui_25K,zheng2024seeact, deng2024mind2web}. To boost model capabilities, we translate query phrases into Chinese and create additional negative samples. We also add in-context visual grounding data, where the task involves locating objects of the same category across multiple images, given a reference object highlighted by a rectangle or ellipse in a reference image.
The data format follows this structure:

\begin{itemize}
\item Prompt: \texttt{<|grounding|>The first image shows <object>.Please identify the object of the same category in the second image.}
\item Response: \texttt{<|ref|><description><|/ref|><|det|>[[x1, y1, x2, y2]]<|/det|>}
\end{itemize}

In this format, \texttt{<|grounding|>}, \texttt{<|ref|>}, \texttt{<|/ref|>}, \texttt{<|det|>}, \texttt{<|/det|>} are special tokens. The \texttt{<object>} placeholder represents phrases like ``an object within the red bounding box'' while \texttt{<description>} is the model's description of the detected object (e.g., ``cat'').

\paragraph{Grounded conversation.}
We construct our grounded conversation data using \cite{ma2025groma,plummer2015flickr30k} to further enhance the model's capabilities established during the pretraining phase.

\paragraph{Text-Only datasets.} 
To maintain the language ability of the model, we also use text-only instruction-tuning datasets~\citep{xu2024magpie, amini2019mathqa, cobbe2021gsm8k, mitra2024orcamath, wei2024magicoder, peng2023instruction, toshniwal2024openmath2, bai2024coig, DatabricksBlog2023DollyV2} during the SFT stage.
\section{Training Methodology}

\subsection{Training Pipelines}
DeepSeek-VL2 is trained through a three-stage pipeline: (1) an initial stage where we train the vision encoder and vision-language adaptor MLP while keeping the language model fixed, using image-text paired data detailed in Section~\ref{data:vl_adaptor}, (2) a pretraining stage where we conduct vision-language pre-training using the data described in Section~\ref{data:vl_pretraining}, and (3) a fine-tuning stage where we perform supervised fine-tuning with the data outlined in Section~\ref{data:vl_sft}.
In both the pretraining and fine-tuning stages, all model parameters, including the vision encoder, vision-language adaptor, and language model, are unlocked and trained simultaneously. Throughout all stages, we emphasize visual understanding capabilities and compute the next token prediction loss exclusively on the text tokens.

\textbf{Vision-Language Alignment.}
Building upon pre-trained language models (DeepSeekMoE 3B/16B/27B), our primary objective is to establish robust connections between visual features and language features. This alignment enables the pre-trained language model to effectively handle visual inputs. Unlike previous approaches~\citep{liu2024visual, lu2024deepseek}, which maintain fixed pretrained vision encoders and language models, we adapt the fixed-resolution vision encoder to accommodate dynamic high-resolution images. In this stage, we optimize both the vision encoder and vision-language adaptor while keeping the language model frozen.

\begin{table}[ht]
\caption{
\textbf{Hyperparameters for training DeepSeek-VL2}. The Step LR Scheduler divides the learning rate by $\sqrt{10}$ at 50\% and 75\% of the total training steps.
}
\centering
\setlength{\tabcolsep}{2pt}
\renewcommand{\arraystretch}{1.2}
\scalebox{0.75}{
\begin{tabular}{l|ccc|ccc|ccc}
\shline
& \multicolumn{3}{c}{DeepSeek-VL2-Tiny} \vline & \multicolumn{3}{c}{DeepSeek-VL2-Small} \vline & \multicolumn{3}{c}{DeepSeek-VL2} \\
\shline
Total parameters~(LLM) & \multicolumn{3}{c}{3B} \vline & \multicolumn{3}{c}{16B} \vline & \multicolumn{3}{c}{27B} \\
Activated parameters~(LLM) & \multicolumn{3}{c}{0.57B} \vline & \multicolumn{3}{c}{2.4B} \vline & \multicolumn{3}{c}{4.1B} \\
Vision Encoder& \multicolumn{3}{c}{SigLIP-SO400M} \vline & \multicolumn{3}{c}{SigLIP-SO400M} \vline & \multicolumn{3}{c}{SigLIP-SO400M} \\
\hline
\textbf{Hyperparameters} & \textbf{Stage 1} & \textbf{Stage 2}         & \textbf{Stage 3} & \textbf{Stage 1} & \textbf{Stage 2}  & \textbf{Stage 3} & \textbf{Stage 1} & \textbf{Stage 2}  & \textbf{Stage 3} \\
\hline
Learning rate         & $5.4\times10^{-4}$   & $5.4\times10^{-4}$ & $3.0\times10^{-5}$& $4.2\times10^{-4}$   & $4.2\times10^{-4}$ & $1.4\times10^{-5}$ & $4.5\times10^{-4}$ & $4.5\times10^{-4}$ & $2\times10^{-5}$ \\
Visual Encoder LR multiplier & 0.1 & 0.1 & 0.1 & 0.1 & 0.1 & 0.1 & 0.1 & 0.1 & 0.1 \\
Fix langauge model & \checkmark & $\times$ & $\times$ & \checkmark & $\times$ & $\times$ & \checkmark & $\times$ & $\times$ \\
LR scheduler          & Cosine   & Step & Constant & Cosine   & Step & Constant & Cosine   & Step & Constant  \\
Weight decay          & 0.1      & 0.1& 0.1& 0.1& 0.1& 0.1 & 0.1& 0.1& 0.1   \\
Gradient clip         & 1.0      & 1.0 & 1.0 & 1.0 & 1.0 & 1.0 & 1.0 & 1.0 & 1.0 \\
Optimizer             & \multicolumn{3}{c}{AdamW($\beta_1=0.9, \beta_2=0.95$)} \vline & \multicolumn{3}{c}{AdamW($\beta_1=0.9, \beta_2=0.95$)} \vline & \multicolumn{3}{c}{AdamW($\beta_1=0.9, \beta_2=0.95$)} \\
BF16 optimizer & $\times$ & $\times$ & $\times$ & $\times$ & $\times$ & $\times$ & \checkmark & \checkmark & \checkmark \\
Aux loss weight & 0.001 & 0.001 & 0.001 & 0.001 & 0.001 & 0.001 & 0.0001 & 0.0001 & 0.0001  \\
Expert bias correction step & - & - & - & - & - & - & 0& 0.001 & 0\\
Training tokens & $2.0$B & $798.5$B & $19.5$B & $2.0$B & $808.9$B & $20.0$B & $2.0$B & $796.5$B & $19.5$B \\
Batch size            & 256      & 2304 & 64 & 256      & 2304 & 64 & 256 & 3360 & 64    \\
Sequence length       & 4096      & 4096 & 4096 & 4096      & 4096 & 4096 & 4096      & 4096 & 4096   \\
Sequence packing      & $\times$ & \checkmark & \checkmark & $\times$ & \checkmark & \checkmark & $\times$ & \checkmark & \checkmark \\
Pipeline parallelism  & $\times$ & \checkmark & \checkmark & \checkmark & \checkmark & \checkmark & \checkmark & \checkmark & \checkmark \\
\shline
\end{tabular}}
\label{tab:hyper}
\end{table}

\textbf{Vision-Language Pre-training.}
After establishing the vision-language alignment in the embedding space, we dedicate the majority of our computational resources to vision-language pre-training. This stage focuses on developing comprehensive joint vision-language knowledge across diverse tasks. We unfreeze all parameters, including the vision encoder, vision-language adaptor MLP, and DeepSeekMoE LLM, to enable full model optimization. Using approximately 800B image-text tokens (Section~\ref{data:vl_pretraining}), this stage significantly enhances the model's multimodal understanding capabilities while maintaining most of its language capabilities.

\textbf{Supervised Fine-Tuning.}
In the final stage, we enhance the pre-trained model's instruction-following and conversational capabilities through supervised fine-tuning. Using our in-house vision-language SFT data, we optimize all parameters while supervising only the answers and special tokens, masking both system and user prompts. To strengthen dialogue comprehension, we combine multimodal data with the pure text dialogue data from DeepSeek-V2~\citep{ds_v2}. This approach ensures robust performance across diverse vision-language tasks, including dense image captioning, general VQA, OCR, table/chart/document/figure understanding, visual-to-code, visual reasoning, visual grounding, and language understanding, etc..

\begin{table}[ht]
\centering
\caption{\textbf{Comparison with state-of-the-art models on OCR-related multimodal benchmarks}. $^\dag$: activated parameters of MoE model.}
\vspace{-2mm}
\renewcommand{\arraystretch}{1.3}
\label{tab:my-table qa_ocr}
\scalebox{0.65}{
\begin{tabular}{c|ccc|ccccc}
\shline
 \multirow{2}{*}{Model} & \multirow{2}{*}{\#Params (LLM)} & \multirow{2}{*}{\#Params (VE)} & \multirow{2}{*}{\#Params (Activated)} & DocVQA & ChartQA & InfoVQA & TextVQA & OCRBench \\
 &  &  &  & (test) & (test) & (test) & (val) & \\ \shline
\multicolumn{9}{c}{\textbf{Closed Model}} \\ \shline
GPT-4V~\cite{openai2023gpt4v} & - & - & - & 87.2 & 78.1 & 75.1 & 78.0 &  645 \\
GPT-4o~\cite{hurst2024gpt} & - &-  & - & 92.8 & 85.7 & 79.2 & 77.4 &  736 \\
Claude 3.5 Sonnet~\cite{anthropic2024claude} &-  & - & - & 95.2 & 90.8 & 74.1 & 74.1 &  788 \\
Gemini-1.5-Pro~\cite{team2024gemini} & - &-  &-  & 93.1 & 87.2 & 80.1 & 78.7 &  754 \\ \shline
\multicolumn{9}{c}{\textbf{Open-source Model~(0.5B - 3B)}} \\ \shline
LLaVA-OV 0.5B~\cite{li2024llava} & 0.5B & 0.4B & 0.9B & 70.0 & 61.4 & 41.8 &  - & - \\
InternVL2-1B~\cite{chen2024internvl2} & - & - & 0.9B & 81.7 & 72.9 & 50.9 & 70.5 &  754 \\
MM 1.5-1B~\cite{zhang2024mm}  & - & - & 1B & 81.0 & 67.2 & 50.5 & 72.5 & 605 \\
\rowcolor{blue!4}
\textbf{DeepSeek-VL2-Tiny} & 0.6B$^\dag$ & 0.4B & 1.0B$^\dag$ & 88.9 & 81.0 & 66.1 & 80.7 & 809 \\ 
MolmoE-1B~\cite{deitke2024molmo} & 1.2B$^\dag$ & 0.3B & 1.5B$^\dag$ & 77.7 & 78.0 & 53.9 & 78.8 &   \\
MiniCPM-V 2.0~\cite{yao2024minicpm} & 2.4B & 0.4B & 2.8B & 71.9 & - & - & 74.1 & 605 \\
InternVL2-2B~\cite{chen2024internvl2} & 1.9B & 0.3B & 2.2B & 86.9 & 76.2 & 58.9 & 73.4 & 784 \\
Qwen2-VL-2B~\cite{wang2024qwen2} & 1.5B & 0.7B & 2.2B & 90.1 & 73.5 & 65.5 & 79.7 &  794 \\
MM 1.5-3B~\cite{zhang2024mm}  & - & - & 3B & 87.7 & 74.2 & 58.5 & 76.5 & 657 \\
\rowcolor{blue!4}
\textbf{DeepSeek-VL2-Small} & 2.4B$^\dag$ & 0.4B & 2.8B$^\dag$ & 92.3 & 84.5 & 75.8 & 83.4 & 834 \\ \shline
\multicolumn{9}{c}{\textbf{Open-source Model~(4B - 13B)}} \\ \shline
Phi-3.5-Vision~\cite{abdin2024phi} & 3.8B & 0.3B & 4.1B & 69.3 & 81.8 & 36.6 & 72.0 &  599 \\
InternVL2-4B~\cite{chen2024internvl2} & 3.8B & 0.3B & 4.1B & 89.2 & 81.5 & 67.0 & 74.4 &  788 \\
Aria-MoE~\cite{li2024aria} & 3.9B$^\dag$ & 0.4B & 4.3B$^\dag$ & 92.6 & 86.4 & - & 81.1 & -\\
MM 1.5-7B~\cite{zhang2024mm}  & - & - & 7B & 88.1 & 78.6 & 59.5 & 76.5 & 635 \\
LLaVA-OV 7B~\cite{li2024llava} & 7.6B & 0.4B & 8.0B & 87.5 & 80.0 & 68.8 & - & -  \\
Molmo-7B-O~\cite{deitke2024molmo} & 7.3B & 0.3B & 7.6B & - & 80.4 & 70.0 & 80.4 & -  \\
MiniCPM-V2.6~\cite{yao2024minicpm} & 7.6B & 0.4B & 8.0B & 90.8 & 82.4 & - & 80.1 & 852 (CoT)  \\
InternVL2-8B~\cite{chen2024internvl2} & 7.7B & 0.3B & 8.0B & 91.6 & 83.3 & 74.8 & 77.4 &  794 \\
Qwen2-VL-7B~\cite{wang2024qwen2} & 7.6B & 0.7B & 8.3B & 94.5 & 83.0 & 76.5 & 84.3 &  845 \\
Pixtral-12B~\cite{agrawal2024pixtral} & 12.0B & 0.4B & 12.4B & 90.7 & 81.8 (CoT) & 50.8 & 75.7 &   \\
DeepSeek-VL 7B~\cite{lu2024deepseek} & 6.9B & 0.4B & 7.3B & - & - & -  & - & 456 \\
\rowcolor{blue!4}
\textbf{DeepSeek-VL2} & 4.1B$^\dag$ & 0.4B & 4.5B$^\dag$ & 93.3 & 86.0 & 78.1 & 84.2 & 811 \\ \shline
\end{tabular}
}
\end{table}

\subsection{Hyperparameters and Infrastructures}

Detailed hyperparameters for DeepSeek-VL2 training are listed in Table~\ref{tab:hyper}. We conducted our training and evaluation using HAI-LLM~\citep{2023HAI-LLM}, an efficient and lightweight platform designed for large models. A significant challenge in our pipeline parallel strategy arose from the vision encoder's unique computational characteristics compared to LLM blocks. As the first component in the model pipeline, the vision encoder requires careful load balancing across GPUs to prevent pipeline bubbles and optimize GPU utilization. To address this, we implemented fine-grained layer division of the vision encoder within our pipeline parallel strategy. Moreover, we perform image tile load balancing across different data parallel ranks during the forward and backward processes to alleviate the imbalance in the number of image tiles caused by the dynamic resolution strategy. Our training process also incorporates tensor parallelism and expert parallelism approaches to achieve the highest efficiency. Since some data batches have only text data while others include image data, we introduce two different pipeline strategies for different kinds of data and switch between these two strategies on demand.
The training of DeepSeek-VL2 was completed in 7/10/14 days using a cluster of 16/33/42 nodes, with each node equipped with 8 NVIDIA A100 GPUs.
\section{Evaluation}
\subsection{Multimodal Performance}

\paragraph{Benchmarks} We perform a holistic evaluation of DeepSeek-VL2 across a collection of commonly used benchmarks, including DocVQA~\citep{mathew2021docvqa}, ChartQA~\citep{masry2022chartqa}, InfoVQA~\footnote{Given that InfoVQA contains images with extreme aspect ratios and excessively large images, we enlarge the candidate resolutions as $C_R = \{(m \cdot 384, n \cdot 384)~\mid~m \in \mathbb{N}, n \in \mathbb{N},1\le m,n,mn\le18\}$ when evaluating.}~\citep{mathew2022infographicvqa}, TextVQA~\citep{singh2019towards}, RealWorldQA~\citep{realworldqa}, OCRBench~\citep{liu2023hidden}, AI2D~\citep{kembhavi2016diagram}, MMMU~\citep{yue2024mmmu}, MMStar~\citep{chen2024we}, MathVista~\citep{lumathvista}, MME~\citep{fu2024mme}, MMBench, MMBench-V1.1~\citep{liu2025mmbench} and MMT-Bench~\citep{yingmmt}. These benchmarks span diverse tasks from document understanding and chart interpretation to real-world problem solving, enabling comprehensive evaluation of our model's capabilities.
To evaluate the grounding capability of our models, we test DeepSeek-VL2 on the RefCOCO, RefCOCO+ and RefCOCOg benchmarks~\citep{kazemzadeh2014referitgame, mao2016generation}.

\begin{table}[ht]
\centering
\caption{\textbf{Comparison with state-of-the-art models on general QA and math-related multimodal benchmarks}. $^\dag$: activated parameters of MoE model. *: evaluated in a different  setting.}
\vspace{-2mm}
\renewcommand{\arraystretch}{1.2}
\label{tab:my-table-reasoning}
\scalebox{0.52}{
\begin{tabular}{c|c|cccccccccc}
\shline
\multirow{2}{*}{Model} & \#Params & MMStar & AI2D & MMMU & MME & MMBench & MMBench & MMBench-V1.1 & MMT-Bench & RealWorldQA & MathVista \\
 & (Activated) & & (test) & (val) &  & (sum) & (en test) & (cn test) &  & &  (testmini) \\
 \shline
\multicolumn{12}{c}{\textbf{Closed Model}} \\ \shline
GPT-4V~\cite{openai2023gpt4v} & - & 56.0 & 89.4 & 63.1 & 1,927 & 81 & 80.2 & 80 & 64.3 & 61.4 & 58.1 \\
GPT-4o~\cite{hurst2024gpt} & - & 63.9 & 94.2 & 69.1 & 2,329 & 83.4 & 82.1 & 82.2 & 65.5 & 75.4 & 63.8 \\
Claude 3.5 Sonnet~\cite{anthropic2024claude} & - & 62.2 & 94.7 & 68.3 & 1,920 & 79.7 & 80.7 & 78.5 & - & 60.1 & 67.7 \\
Gemini-1.5-Pro~\cite{team2024gemini} & - & - & 94.4 & 62.2 & - & - & - & - & 64.5 & 70.4 & 63.9 \\  \shline
\multicolumn{12}{c}{\textbf{Open-source Model~(0.5B - 3B)}} \\ \shline
LLaVA-OV 0.5B~\cite{li2024llava} & 0.9B & 37.7 & 57.1 & 31.4 & 1,478 & 61.6 & 55.5 & 59.6 & - & 55.6 & 34.8\\
InternVL2-1B~\cite{chen2024internvl2} & 0.9B & 45.7 & 64.1 & 35.4 & 1,794 & 65.4 & 60.7 & 61.6 & 49.5 & 50.3 & 37.7 \\
MM 1.5-1B~\cite{zhang2024mm} & 1B & - & 59.3 & 35.8 & 1,611 & - & - & - & - & 53.3 & 37.2 \\
\rowcolor{blue!4}
\textbf{DeepSeek-VL2-Tiny} & 1.0B$^\dag$ & 45.9 & 71.6 & 40.7 & 1,915 & 73.3 & 69.2 & 68.3 & 53.2 & 64.2 & 53.6 \\
MolmoE-1B~\cite{deitke2024molmo} & 1.5B$^\dag$ & - & 86.4* & 34.9 & - & - & - & - & - & 60.4 & 34\\
MiniCPM-V 2.0~\cite{yao2024minicpm} & 2.8B & - & - & 38.2 & 1,809 & 69.6 & 68.1 & - & - & - & 38.7 \\ 
InternVL2-2B~\cite{chen2024internvl2}   & 2.2B & 49.8 & 74.1 & 36.3 & 1,877 & 73.2 & 70.9 & 69.6 & 50.4 & 57.3 & 46.3 \\
Qwen2-VL-2B~\cite{wang2024qwen2} & 2.2B & 48 & 74.4 & 41.1 & 1,872 & 74.9 & 73.5 & 72.2 & 54.5 & 62.9 & 47.8 \\
MM 1.5-3B~\cite{zhang2024mm} & 3B & - & 65.7 & 37.1 & 1,798 & - & - & - & - & 56.9 & 44.4 \\
\rowcolor{blue!4}
\textbf{DeepSeek-VL2-Small} & 2.8B$^\dag$ & 57.0 & 80.0 & 48.0 & 2,123 & 82.3 & 80.3 & 79.3 & 62.9 & 65.4 & 60.7\\ \shline
\multicolumn{12}{c}{\textbf{Open-source Model~(4B - 13B)}} \\ \shline
Phi-3.5-Vision~\cite{abdin2024phi} & 4.1B & 47.5 & 78.1 & 43 & - & 76 & 66.1 & 72.1 & 53.6 & 53.6 & 43.9 \\
InternVL2-4B~\cite{chen2024internvl2} & 4.1B & 54.3 & 78.9 & 47.9 & 2,060 & 78.6 & 73.9 & 75.8 & 55.7 & 60.7 & 58.6\\
Aria-MoE~\cite{li2024aria} & 4.3B$^\dag$ & - & - & 54.9 & - & - & - & - & - & - & 66.1\\
MM 1.5-7B~\cite{zhang2024mm} & 7B & - & 72.2 & 41.8 & 1,861 & - & - & - & - & 62.5 & 47.6\\
LLaVA-OV 7B~\cite{li2024llava} & 8.0B & - & 81.4 & 48.8 & 1,998 & 80.8 & - & - & - & 66.3 & 63.2 \\
Molmo-7B-O~\cite{deitke2024molmo} & 7.6B$^\dag$ & - & 90.7* & 39.3 &  - & - & - & - & - & 67.5 & 44.5\\
MiniCPM-V2.6~\cite{yao2024minicpm} & 8.0B & 57.5 & 82.1 & 49.8 (CoT) & 2,348 (CoT) & 81.5 & 79.3 & 78.0 & 60.8 & 65.0 & 60.6 \\
InternVL2-8B~\cite{chen2024internvl2} & 8.0B & 61.5 & 83.8 & 51.8 & 2,210 & 81.7 & 81.2 & 79.4 & 60.0 & 64.4 & 58.3 \\
Qwen2-VL-7B~\cite{wang2024qwen2} & 8.3B & 60.7 & 83 & 54.1 & 2,327 & 83 & 80.5 & 80.7 & 63.7 & 70.1 & 58.2 \\
Pixtral-12B~\cite{agrawal2024pixtral} & 12.4B & - & - & 52.5 (CoT) & - & - & - & - & - & 65.4 & 58 (CoT) \\
DeepSeek-VL 7B~\cite{lu2024deepseek} & 7.3B & - & - & 36.6 & - & 73.2 & - & - & - & - & -\\
\rowcolor{blue!4}
\textbf{DeepSeek-VL2} & 4.5B$^\dag$ & 61.3 & 81.4 & 51.1 & 2,253 & 83.1 & 79.6 & 79.2 & 63.6 & 68.4 & 62.8\\ \shline
\end{tabular}
}
\end{table}
\vspace{-2mm}

\begin{table}[t]
\centering
\caption{\textbf{Comparison with state-of-the-art models on visual grounding benchmarks}. Our models of different sizes have all achieved the best results among MLLMs with similar sizes.}
\renewcommand{\arraystretch}{1.2}
\label{tab:my-table grounding}
\setlength{\tabcolsep}{16pt}
\scalebox{0.68}{
\begin{tabular}{c|ccc|ccc|cc}
\shline
\multirow{2}{*}{Model} & \multicolumn{3}{c}{RefCOCO} \vline & \multicolumn{3}{c}{RefCOCO+} \vline & \multicolumn{2}{c}{RefCOCOg} \\
~ & val & testA & testB & val & testA & testB & val & test \\
\shline
\multicolumn{9}{c}{\textbf{Vision Model}} \\
\shline
Grounding DINO-Tiny~\cite{liu2025grounding} & 89.2 & 91.9 & 86.0 & 81.1 & 87.4 & 74.7 & 85.2 & 84.9 \\
Grounding DINO-Largey~\cite{liu2025grounding} & 90.6 & 93.2 & 88.2 & 82.8 & 89.0 & 75.9 & 86.1 & 87.0 \\
UNINEXT-H~\cite{lin2023uninext} & 92.6 & 94.3 & 91.5 & 85.2 & 89.6 & 79.8 & 88.7 & 89.4 \\
\shline
\multicolumn{9}{c}{\textbf{VLM + Task-Specific Fine-Tuning}} \\ 
\shline
ONE-PEACE~\cite{wang2023one} & 92.6 & 94.2 & 89.3 & 88.8 & 92.2 & 83.2 & 89.2 & 89.3 \\
mPLUG-2~\cite{xu2023mplug} & 90.1 & 92.8 & 86.1 & - & - & 86.1 & 84.7 & 85.1 \\
Florence-2-B~\cite{xiao2024florence} & 92.6 & 94.8 & 91.5 & 86.8 & 91.7 & 82.2 & 89.8 & 82.2 \\
Florence-2-L~\cite{xiao2024florence} & 93.4 & 95.3 & 92.0 & 88.3 & 92.9 & 83.6 & 91.2 & 91.7 \\
\shline
\multicolumn{9}{c}{\textbf{Open-source VLM~(0.5B - 3B)}} \\ \shline
InternVL2-1B~\cite{chen2024internvl2} & 83.6 & 88.7 & 79.8 & 76.0 & 83.6 & 67.7 & 80.2 & 79.9 \\
\rowcolor{blue!4}
\textbf{DeepSeek-VL2-Tiny} & 84.7 & 87.8 & 78.4 & 75.9 & 83.9 & 67.4 & 73.8 & 83.9 \\
InternVL2‑2B~\cite{chen2024internvl2} & 82.3 & 88.2 & 75.9 & 73.5 & 82.8 & 63.3 & 77.6 & 78.3 \\
\rowcolor{blue!4}
\textbf{DeepSeek-VL2-Small} & 93.9 & 95.3 & 91.3 & 89.4 & 92.9 & 84.8 & 92.6 & 92.6\\ 
\shline
\multicolumn{9}{c}{\textbf{Open-source VLM~(4B - 9B)}} \\ \shline
Shikra-7B~\cite{chen2023shikra} & 87.0 & 90.6 & 80.2 & 81.6 & 87.4 & 72.1 & 82.3 & 82.2 \\
TextHawk2-7B~\cite{yu2024texthawk2} & 91.9 & 93.0 & 87.6 & 86.2 & 90.0 & 80.4 & 88.2 & 88.1 \\
Ferret-v2-7B~\cite{zhang2024ferret} & 92.8 & 94.7 & 88.7 & 87.4 & 92.8 & 79.3 & 89.4 & 89.3 \\
InternVL2-8B~\cite{chen2024internvl2} & 87.1 & 91.1 & 80.7 & 79.8 & 87.9 & 71.4 & 82.7 & 82.7 \\
MM1.5-7B~\cite{zhang2024mm} & - & 92.5 & 86.7 & - & 88.7 & 77.8 &  - & 87.1 \\
Qwen2-VL-7B~\cite{wang2024qwen2} & 91.7 & 93.6 & 87.3 & 85.8 & 90.5 & 79.5 & 87.3 & 87.8 \\
\rowcolor{blue!4}
\textbf{DeepSeek-VL2} & 95.1 & 96.7 & 92.7 & 91.2 & 94.9 & 87.4 & 92.8 & 92.9 \\ \shline
\end{tabular}
}
\end{table}
\vspace{-2mm}

\paragraph{Comparison with the state-of-the-arts} On the multimodal understanding benchmarks, we compare DeepSeek-VL2 with state-of-the-art models, including LLaVA-OV~\citep{li2024llava}, InternVL2~\citep{chen2023internvl}, DeepSeek-VL~\citep{lu2024deepseek}, Qwen2-VL~\citep{wang2024qwen2}, Phi-3.5-Vision~\citep{abdin2024phi}, Molmo~\citep{deitke2024molmo}, Pixtral~\citep{agrawal2024pixtral}, MM1.5~\cite{zhang2024mm} and Aria-MoE~\citep{li2024aria}. The results are reported in Table~\ref{tab:my-table qa_ocr} and~\ref{tab:my-table-reasoning}. 
Benefited from our MoE architecture, DeepSeek-VL2 achieves similar or better performance with fewer activated parameters.
On the grounding benchmarks, we compare DeepSeek-VL2 with Groudning DINO~\cite{liu2025grounding}, UNINEXT~\cite{lin2023uninext}, ONE-PEACE~\cite{wang2023one}, mPLUG-2~\cite{xu2023mplug}, Florence-2~\cite{xiao2024florence}, InternVL2~\cite{chen2024internvl2}, Shikra~\cite{chen2023shikra}, TextHawk2~\cite{yu2024texthawk2}, Ferret-v2~\cite{zhang2024ferret}, MM1.5~\citep{zhang2024mm} and Qwen2~\citep{wang2024qwen2}. Our models outperforms the other VLMs at similar scales.

\begin{figure}[!htp]
	\centering
	\includegraphics[width=0.9\linewidth]{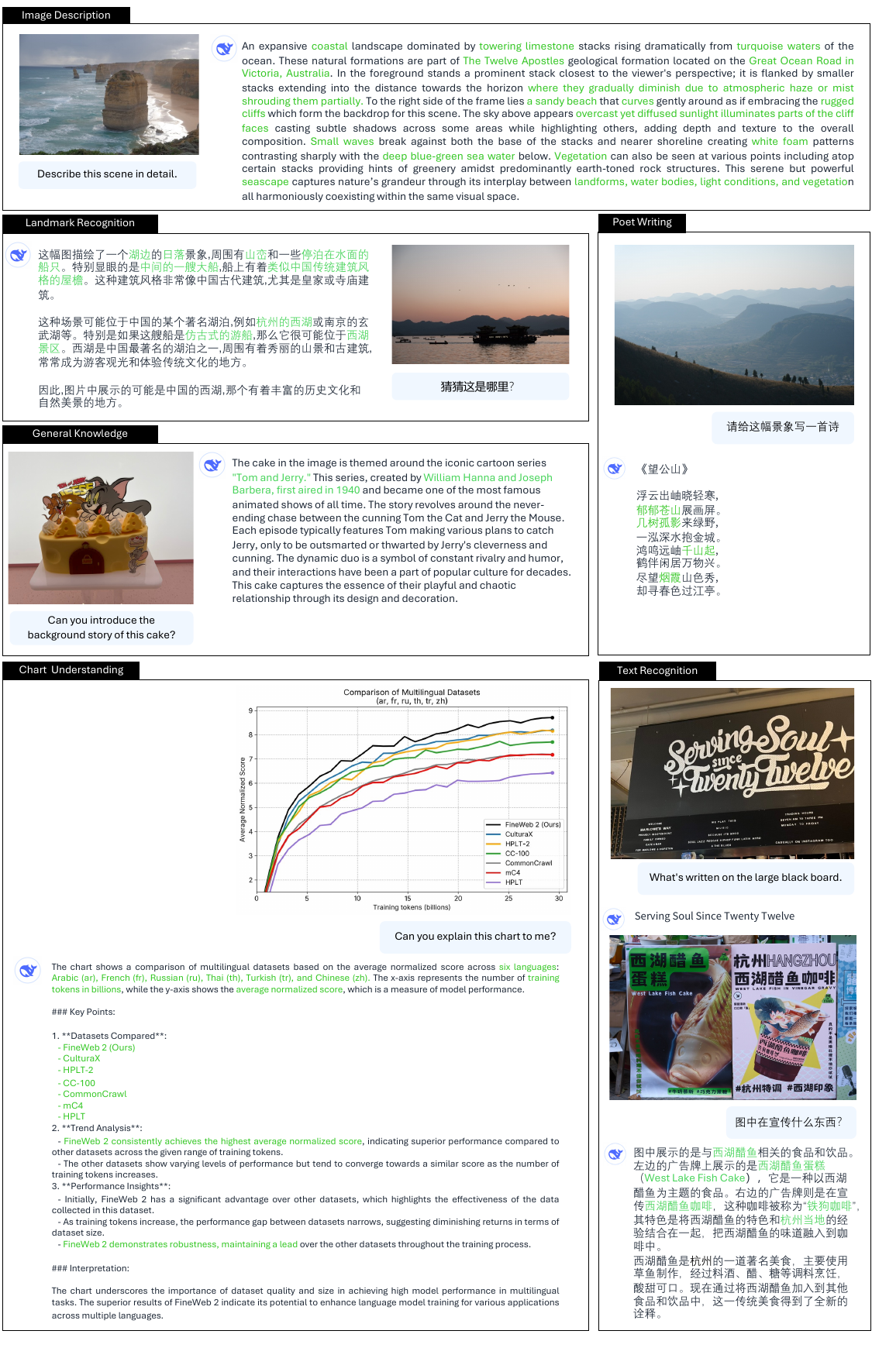}
	\caption{\textbf{General question answering capability of \codename}. Our model has multi-faceted capabilities, including recognition of landmarks, image-based poetry composition, answering questions about general knowledge, understanding charts, recognizing text, and more.}
	\label{fig:cases_general_qa}
\end{figure}

\begin{figure}[!htp]
    \centering
    \includegraphics[width=\linewidth]{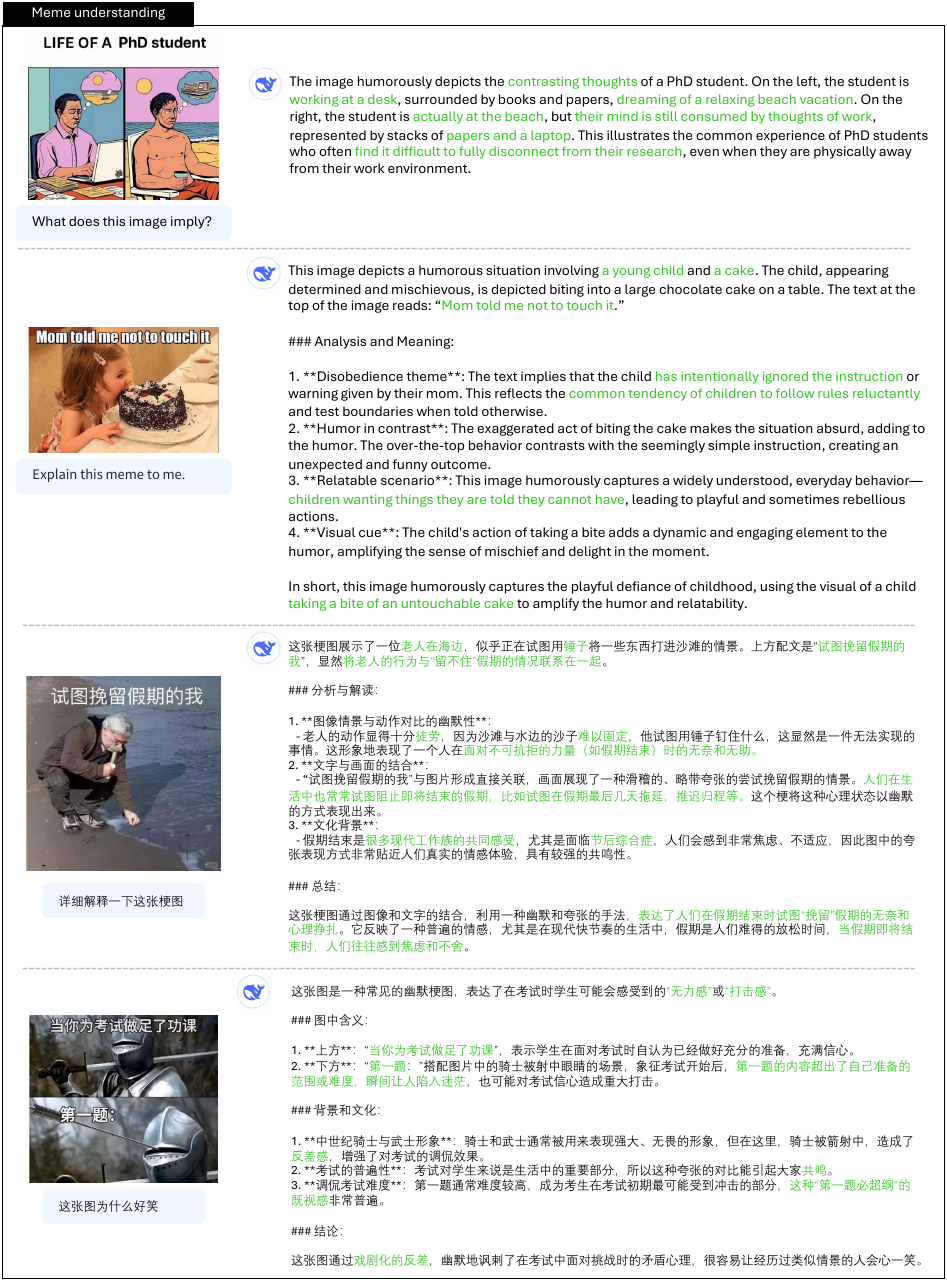}
    \caption{\textbf{Meme understanding capability of \codename}. Our model can understand the humor in memes and provide explanations.}
    \label{fig:cases_meme_understanding}
\end{figure}

\begin{figure}[!htp]
    \centering
    \includegraphics[width=\linewidth]{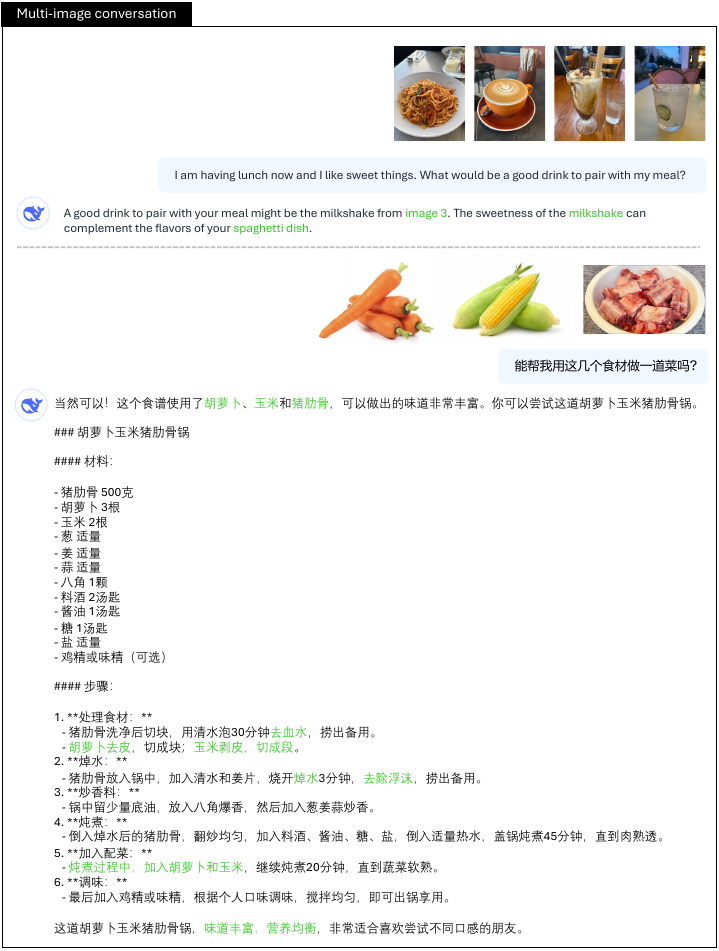}
    \caption{\textbf{Multi-image conversation capability of \codename}. Our model can understand the relationships between multiple images and do some reasoning based on them.}
    \label{fig:cases_multi_image_conversation}
\end{figure}

\begin{figure}[!htp]
	\centering
	\includegraphics[width=\linewidth]{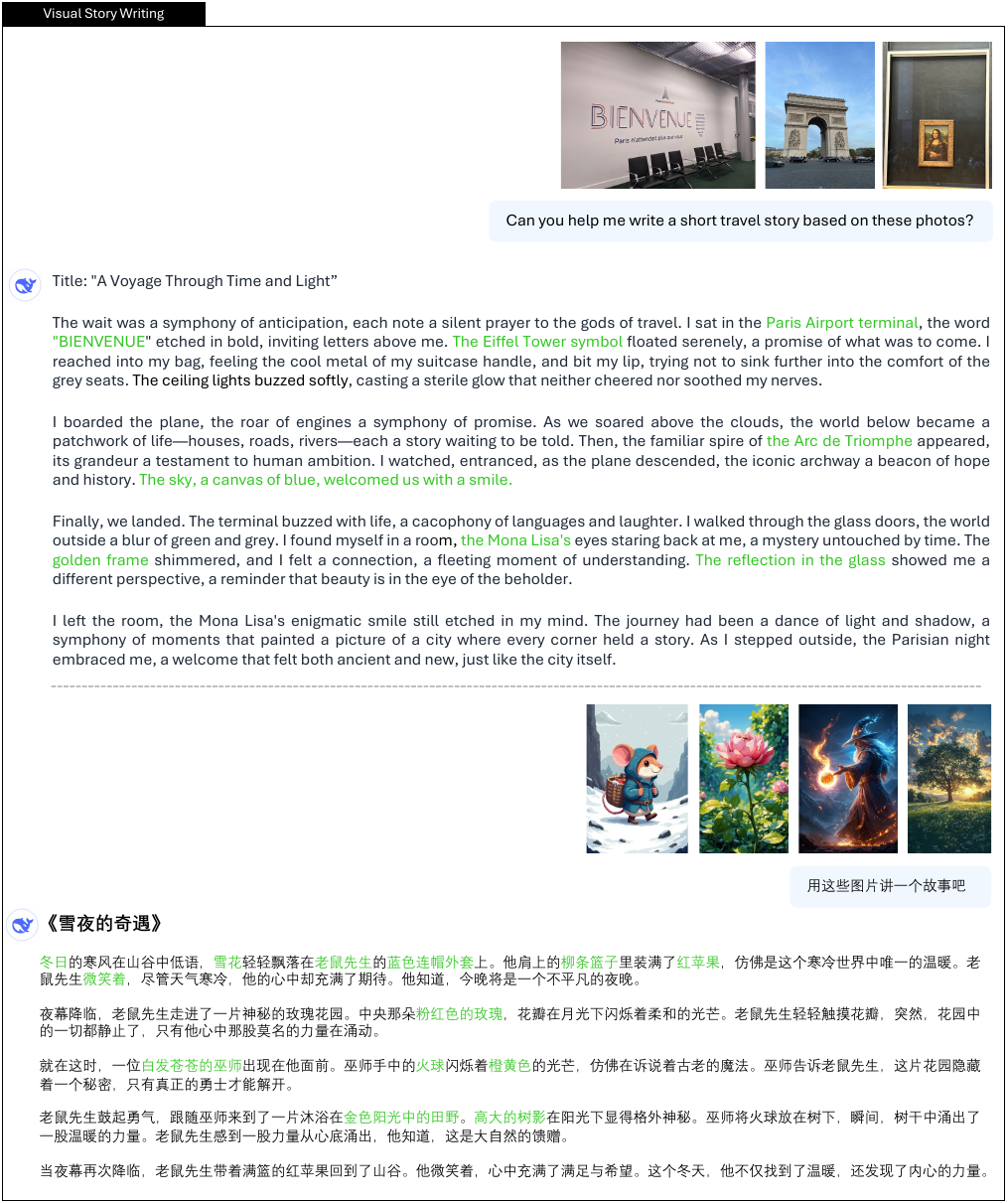}
	\caption{\textbf{Visual storytelling capability of \codename}. Our model can accept multiple images as input and narrate a story in either Chinese or English based on the images.}
	\label{fig:cases_visual_story_telling}
\end{figure}

\subsection{Qualitative Study}
In this section, we demonstrate different capabilities of DeepSeek-VL2, ranging from general question answering to visual storytelling and visual grounding.

\paragraph{General visual question answering.}
Benefited from our new VL pretraining dataset and diverse SFT data. DeepSeek-VL2 demonstrated significantly improved ability on general visual question answering, as shown in Figure~\ref{fig:cases_general_qa}. Overall, this model excels at dense image description and it is able to recognize common landmarks, general visual knowledge, and rich-texts in both English and Chinese. It also performs favorably on chart understanding with accurate attributes recognition. Furthermore, we show the improved meme understanding of DeepSeek-VL2 in Figure~\ref{fig:cases_meme_understanding}, where it can describe the correct context and explain the humor with meaningful cultural background.

\paragraph{Multi-image conversation.}
DeepSeek-VL2 demonstrated improved ability on multi-image conversation, as shown in Figure~\ref{fig:cases_multi_image_conversation}. Our model can analyze the associations and differences among multiple images, while also enabling simple reasoning by integrating the content of several images. For example, it can think about how to prepare a dish based on images of certain ingredients.

\paragraph{Visual storytelling.} 
In Figure~\ref{fig:cases_visual_story_telling}, we show DeepSeek-VL2 is able to write a creative story given a few images. The story writing is backed by its strong general visual capabilities such as landmark recognition and OCR, as highlight in green texts. In addition, since the story writing ability is originally from the text-only DeepSeek Chat model, which is already aligned with good safety, we do not observe significant harmful and NSFW output from DeepSeek-VL2 during our internal testing. However, it is worth noting that creative storytelling in real-world scenarios demands more diverse genres (\eg, horror, comedy, action) and varied plot types (\eg, happy or tragic endings), which may inherently conflict with the safety requirements in LLM/VLM research. We aim to explore solutions to broaden the scope of storytelling while considering these challenges.

\begin{figure}[!htp]
	\centering
	\includegraphics[width=\linewidth]{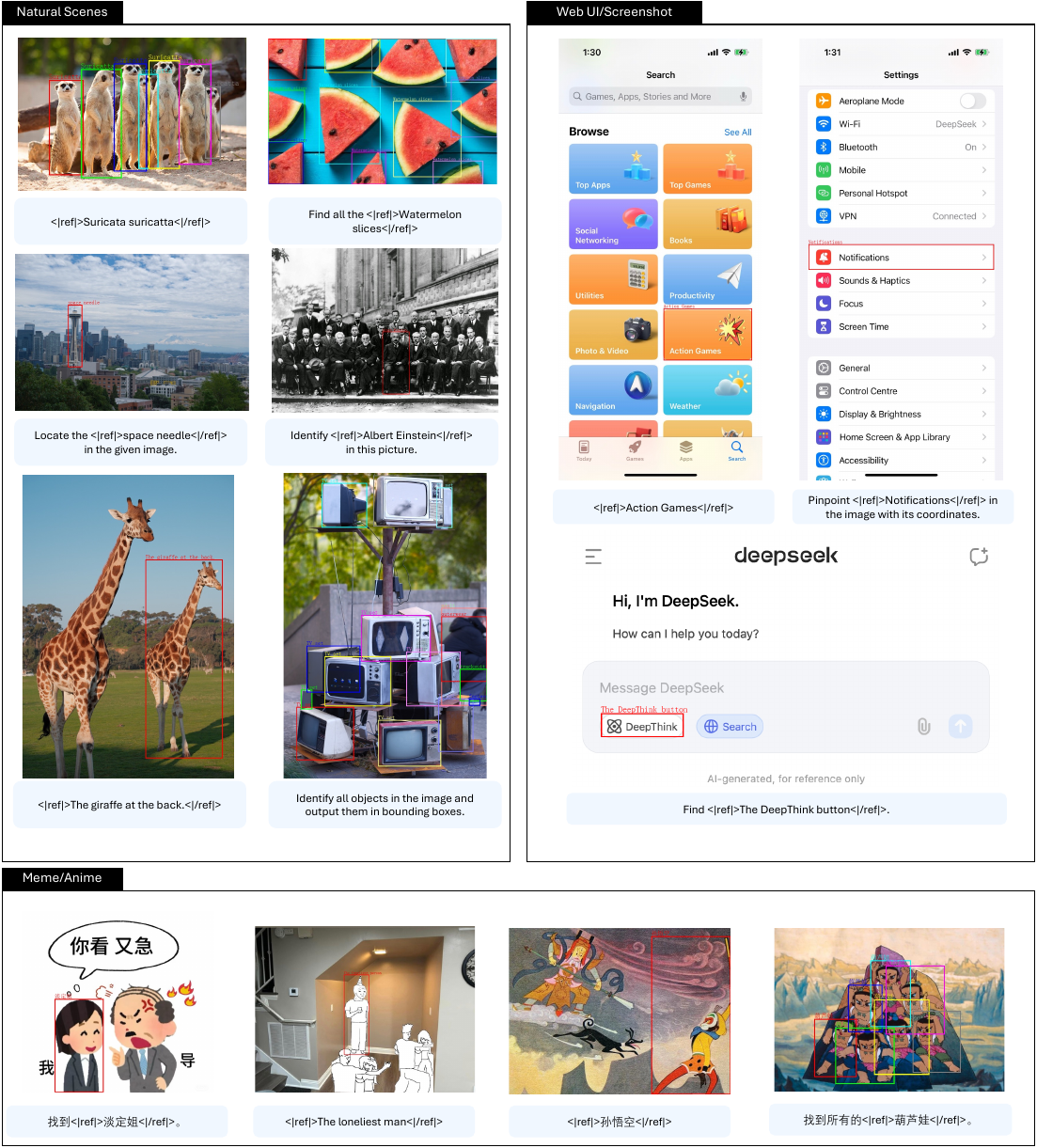}
	\caption{\textbf{Visual grounding ability of \codename}. Our model can locate objects based on their category names, descriptions, or some abstract concepts. }
	\label{fig:cases_visual_grounding_1}
\end{figure}

\begin{figure}[!htp]
	\centering
	\includegraphics[width=\linewidth]{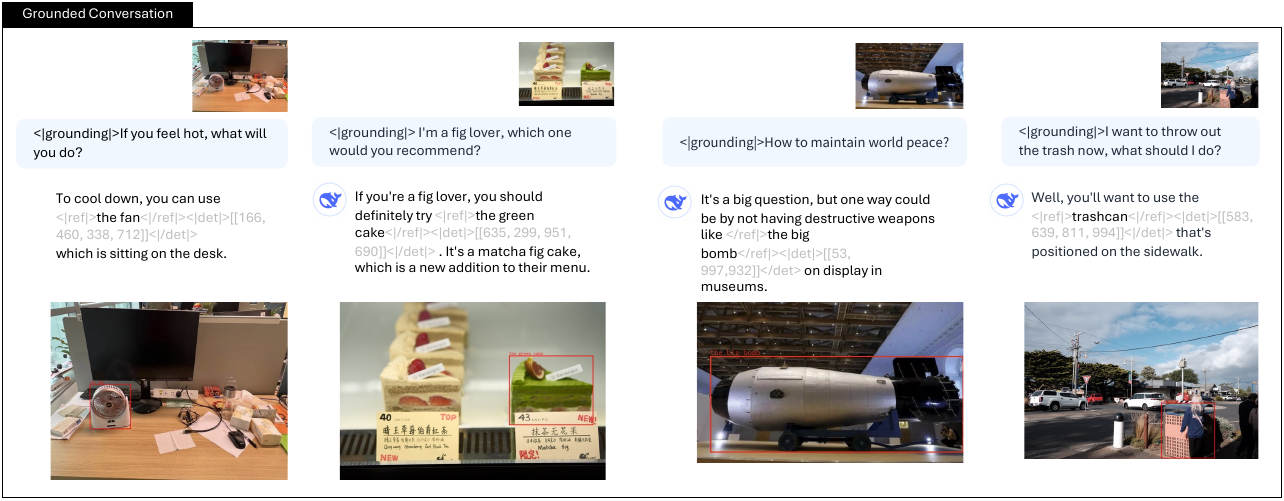}
	\caption{\textbf{Grounded conversation with \codename}. Our model can perform reasoning on images while identifying the locations of relevant objects, thereby enabling the possibility of interacting with the real world.}
	\label{fig:cases_visual_grounded_conversation}
\end{figure}

\begin{figure}[!htp]
	\centering
	\includegraphics[width=\linewidth]{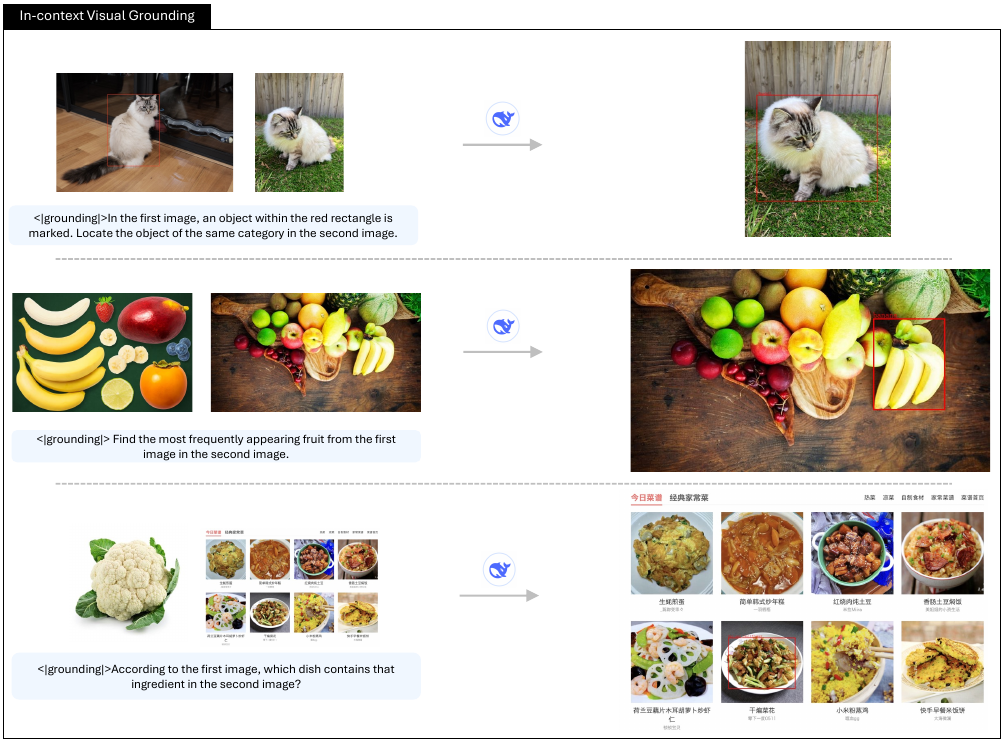}
	\caption{\textbf{In-context visual grounding with DeepSeek-VL2}. Given one image, either with or without visual prompts, DeepSeek-VL2  is able to find relevant objects in another image.}
	\label{fig:cases_in_context_grounding}
\end{figure}

\paragraph{Visual grounding.} 
Visual grounding is a new ability we bring to \codename. In Figure~\ref{fig:cases_visual_grounding_1}, we show the general grounding ability of \codename. 
Interestingly, although the majority of images in our training set come from natural scenes, and the referring expressions are object category names or specific descriptions of objects, we find that the model is capable of generalizing to other scenarios (such as memes and animes), and has the ability to recognize certain celebrities and abstract concepts.
Furthermore, we show \codename~has in-context visual grounding ability in Figure~\ref{fig:cases_in_context_grounding}. Given the first image, where an object is referred by the visual prompt, the model is able to locate the object of the same category in the second image. We also observe that the model has exhibited emergent abilities. Given an image and textual descriptions, the model can combine the information from the image and the text to identify the corresponding object in a second image. Examples are listed in the second and the third rows in Figure~\ref{fig:cases_in_context_grounding}.

\paragraph{Grounding conversation.} 
With the special token \texttt{<|grounding|>}, \codename~can unleash its ability of grounded conversation, where it can refer to the key objects with accurate locations in its response, as demonstrated in Figure~\ref{fig:cases_visual_grounded_conversation}. This enables the model to interact better with the real world, thereby creating opportunities to play a greater role in fields such as embodied AI and computer/phone agents.

\section{Conclusion}

In this technical report, we introduce DeepSeek-VL2, an enhanced version of MoE-based Vision-Language Models, available in scales of 3B, 16B, and 27B parameters in total, with corresponding activated parameters of 1.0B, 2.8B, and 4.5B. This configuration facilitates efficient computational consumption during both training and inference stages. Notably, our 3B, 16B and 27B models can be deployed on a single GPU with 10 GB, 40GB and 80GB memory respectively. We employ a dynamic tiling vision encoding strategy to efficiently process high-resolution images with various aspect ratios. By making codes and pre-trained models publicly available, we aim to stimulate further advancements and applications at the intersection of vision and language.

\paragraph{Limitations and Future Work}
While \codename~demonstrates strong capabilities across various tasks, there are several areas for future improvements. Currently, DeepSeek-VL2's context window only allows for a few images per chat session. We plan to extend the context window in our next version to enable richer multi-image interactions. Moreover, like other current VLMs, the model occasionally faces challenges with blurry images or unseen objects, presenting opportunities for improved robustness in future versions. Finally, while \codename~excels in visual perception and recognition tasks, we aim to strengthen its reasoning capabilities. These identified areas guide our ongoing research directions as we continue to advance the model's capabilities.

\newpage

\bibliography{main}

\setcounter{figure}{0}
\makeatletter 
\renewcommand{\thefigure}{A\@arabic\c@figure}
\makeatother

\setcounter{table}{0}
\makeatletter 
\renewcommand{\thetable}{A\@arabic\c@table}
\makeatother

\end{CJK*}
\end{document}